# Low-Rank Robust Online Distance/Similarity Learning based on the Rescaled Hinge Loss

Davood Zabihzadeh*[1], Amar Tuama[2], Ali Karami-Mollaee[3]

[1]Computer Department, Engineering Faculty, Sabzevar University of New Technology, Sabzevar, IRAN

[2]Computer Department, Engineering Faculty, Ferdowsi University of Mashhad, Mashhad, IRAN

[3]Electrical and Computer Engineering Faculty, Hakim Sabzevari University, Sabzevar, Iran

* Corresponding Author

d.zabihzadeh@mail.um.ac.ir, amar.tuama@mail.um.ac.ir, karami@hsu.ac.ir

## Abstract

An important challenge in metric learning is scalability to both size and dimension of input data. Online metric learning algorithms are proposed to address this challenge. Existing methods are commonly based on (Passive/Aggressive) PA approach. Hence, they can rapidly process large volumes of data with an adaptive learning rate. However, these algorithms are based on the Hinge loss and so are not robust against outliers and label noise. Also, existing online methods usually assume training triplets or pairwise constraints are exist in advance. However, many datasets in real-world applications are in the form of input data and their associated labels. We address these challenges by formulating the online Distance/Similarity learning problem with the robust Rescaled hinge loss function (Xu, Cao et al. 2017). The proposed model is rather general and can be applied to any PA-based online Distance/Similarity algorithm. Also, we develop an efficient robust one-pass triplet construction algorithm. Finally, to provide scalability in high dimensional DML environments, the low-rank version of the proposed methods is presented that not only reduces the computational cost significantly but also keeps the predictive performance of the learned metrics. Also, it provides a straightforward extension of our methods for deep Distance/Similarity learning. We conduct several experiments on datasets from various applications. The results confirm that the proposed methods significantly outperform state-of-the-art online DML methods in the presence of label noise and outliers by a large margin.

## 1. Introduction

The performance of many machine learning and data mining algorithms depends on the metric used to compute Distance/Similarity between data. Generic Distance/Similarity measures such as Euclidean or Cosine similarity in input space often fails to discriminate different classes or clusters of data. Therefore, learning an optimal Distance/Similarity function is actively studied in the last decade.

Distance Metric Learning (DML) methods aim to bring semantically similar data items together while keeping dissimilar ones at a distance. One important challenge for DML algorithms is scalability to both the size and dimension of input data (Bellet, Habrard et al. 2014). For processing massive volumes of data generated in today's applications, online Distance/Similarity methods are proposed.

Many of these algorithms are based on (Passive/Aggressive) PA (Crammer, Dekel et al. 2006) approach (Chechik, Sharma et al. 2010, Xia, Hoi et al. 2014, Wu, Hoi et al. 2016, Zhong, Zheng et al. 2017, Hamdan, Zabihzadeh et al. 2018, Li, Gao et al. 2018, Rasheed, Zabihzadeh et al. 2020) . The main advantages of PA-based methods are closed-form solution and adaptive learning rate leading to a high convergence rate. However, since these algorithms are based on the Hinge loss, they are not robust against outliers and label noise data. Nowadays many modern datasets are collected from the Internet using crowdsourcing or similar techniques. Hence, examples with wrong labels are usual in these datasets that can considerably degrade the performance of existing online DML methods. We address this challenge by formulating the online Distance/Similarity learning task using the robust Rescaled hinge loss function (Xu, Cao et al. 2017). The proposed model is rather general, and we can easily apply it to any existing PA-based methods. It significantly improves the robustness of the existing methods in the presence of label noise without increasing their computational complexity.

Most DML algorithms learn the metric from pair or triplet side information defined as:

$$S = \{(x_i, x_j) \mid x_i \text{ and } x_j \text{ are similar}\}$$
$$D = \{(x_i, x_j) \mid x_i \text{ and } x_j \text{ are dissimilar}\}$$
$$T = \{(x_i, x_i^+, x_i^-) \mid x_i \text{ should be more closer to } x_i^+ \text{ than to } x_i^- \}$$

Existing online methods (Chechik, Sharma et al. 2010, Xia, Hoi et al. 2014, Wu, Hoi et al. 2016, Zhong, Zheng et al. 2017, Hamdan, Zabihzadeh et al. 2018, Li, Gao et al. 2018, Rasheed, Zabihzadeh et al. 2020) usually assume training triplets or pairwise constrains are exist in advance but many datasets in real world applications are not in this format. Also, available batch triplet construction algorithms are very time consuming and often require computing pairwise distances between data items. Thus, these are not applicable for online tasks. We tackle this challenge by developing an efficient robust one-pass triplet construction algorithm.

Another important challenge in online DML applications especially in the field of machine vision is the high dimension of the input data. Many existing methods learn Mahalanobis distance (Xia, Hoi et al. 2014, Zhong, Zheng et al. 2017, Hamdan, Zabihzadeh et al. 2018, Rasheed, Zabihzadeh et al. 2020) or bilinear similarity (Chechik, Sharma et al. 2010, Xia, Hoi et al. 2014) which require $O(d^2)$ parameters ($d$ indicate the data dimension). Therefore, these methods are infeasible in high dimensional environments. We develop the low-rank versions of the proposed methods that reduce the computational cost significantly but also keep the predictive performance of the learned measure. Also, one can easily replace the low-rank projection matrix in the proposed methods with a nonlinear deep neural network model.

Therefore, extending our methods for online deep Distance/Similarity learning is straightforward.

The rest of the paper is organized as follows: Section 2 reviews related works. In Section 3, we present the formulation of the online Distance/Similarity learning problem using the Rescaled Hinge loss as well as the development of the proposed algorithms. Experiments conducted to evaluate the proposed methods are discussed in Section 4. Finally, Section 5 concludes with remarks and recommendations for future work.

## 2. Related Work

Most existing online Distance/Similarity learning methods learn Mahalanobis distance (Wu, Hoi et al. 2016, Zhong, Zheng et al. 2017, Hamdan, Zabihzadeh et al. 2018, Li, Gao et al. 2018, Rasheed, Zabihzadeh et al. 2020) or bilinear similarity (Chechik, Sharma et al. 2010, Xia, Hoi et al. 2014).

Most research in online metric learning is dedicated to learning Mahalanobis distance. Although some newer and more generic measures such as (Lin, Wang et al. 2017, Zabihzadeh, Monsefi et al. 2018) are also presented. Mahalanobis-based methods learn a matrix $M \succcurlyeq 0$ given by:

$$d_M(x_i, x_j)^2 = (x_i - x_j)^T M (x_i - x_j) \qquad (1)$$

Since the matrix $M \succcurlyeq 0$, it can be decomposed as $M = LL^T$ where $L \in \mathbb{R}^{d \times r}$ and $r = rank(M)$. Therefore, Mahalanobis distance learning is equivalent to find a linear transformation $L$ in the input space. On the other hand, bilinear similarity-based methods learn a similarity matrix $M$ given by:

$$S_M(x_i, x_j)^2 = x_i^T M x_j \qquad (2)$$

The optimization problem of both Mahalanobis and bilinear methods is formulated based on the PA approach. We can show this optimization problem in the following general form:

$$M_{t+1} = \arg\min_M reg(M, M_t) + C\xi$$
$$\text{subject to} \quad l(M, S_t) \leq \xi, \qquad \xi \geq 0, \qquad M \succcurlyeq 0 \qquad (3)$$

where $M_t$ is the current Distance/Similarity matrix at time step t. $reg(M, M_t)$ is the regularization term. $S_t$ denotes the input constraint arrived at time t. $S_t$ is often in the form of triplet $(p_t, p_t^+, p_t^-)$, and $l(M, S_t)$ indicates the margin-based Hinge loss function. In distance-based methods, the Hinge loss function is defined as:

$$l(M, (p_t, p_t^+, p_t^-)) = \max\{0, 1 + d_M(p_t, p_t^+)^2 - d_M(p_t, p_t^-)^2\} \qquad (4)$$

On the other hand, it is defined in similarity-based methods as:

$$l(\boldsymbol{M}, (\boldsymbol{p}_t, \boldsymbol{p}_t^+, \boldsymbol{p}_t^-)) = \max\{0, 1 - S_M(\boldsymbol{p}_t, \boldsymbol{p}_t^+)^2 + S_M(\boldsymbol{p}_t, \boldsymbol{p}_t^-)^2\} \quad (5)$$

OASIS[1] (Chechik, Sharma et al. 2010) is a popular bilinear similarity learning method that uses Frobenius norm as a regularization term, i.e. $reg(\boldsymbol{M}, \boldsymbol{M}_t) = \frac{1}{2}\|\boldsymbol{M} - \boldsymbol{M}_t\|_F^2$. OASIS eliminates p.s.d (positive semi-definite) constraint for scalability reason. However, this property is very useful to produce a low-rank metric as well as to prevent overfitting. This work is extended in OMKS[2] (Xia, Hoi et al. 2014) for multi-modal data. OMKS learns a separate linear similarity operator for each source of data in the feature space induced by an RKHS kernel. The final similarity function is a weighted average of these similarity measures. The weights are updated using Hedge (Freund and Schapire 1997) in an online manner.

OMDML[3] (Wu, Hoi et al. 2016) is similar to OMKS, but instead of bilinear similarity, it learns Mahalanobis distance for each source of data. To enforce p.s.d constraint per metric, OMKS uses full Eigen value decomposition which involves $O(d^3)$ operations. Therefore, this method is infeasible for high-dimensional DML tasks. To address this problem, LSMDML [4] (Rasheed, Zabihzadeh et al. 2020) utilizes DRP (Dual Random Projection) (Qian, Jin et al. 2015) in an online multi-modal environment to enforce p.s.d constraint per metric. DRP considerably decreases the time of the metric learning process in high-dimensional datasets while preserves the performance of learned metric. Also, LSMDML combines the learned metrics using a novel PA-based method which leads to a better convergence rate in comparison with the traditional Hedge algorithm.

SLMOML[5] (Zhong, Zheng et al. 2017) is the online version of the seminal ITML[6] (Davis, Kulis et al. 2007) method. It uses the *logdet* regularization term which automatically enforces p.s.d constraint at each time step. However, it has a low convergence rate and requires $O(d^2)$ parameters.

In (Hamdan, Zabihzadeh et al. 2018) a large-scale local online Distance/Similarity framework is presented. It learns multiple metrics for the task at hand, one metric per class in the dataset. Each metric in this framework consists of a global and a local component learned simultaneously in online fashion. It can find a nonlinear projection by learning multiple local metrics. Also, adding the global component to local metrics shares discriminating information between them and efficiently reduces the overfitting problem. Also, this framework utilizes DRP (Dual Random Projection) to achieve scalability respect to the data dimension.

---

[1] Online Algorithm for Scalable Image Similarity

[2] Online Multiple Kernel Similarity Learning

[3] Online Multi-Modal Distance Metric Learning

[4] Large-Scale Multi-modal Distance Metric Learning

[5] Scalable Large Margin Online Metric Learning

[6] Information-Theoretic Metric Learning

OPML[1] (Li, Gao et al. 2018) is an online DML method which learns projection matrix $L$ (see equation (1)) directly, so it does not require imposing the p.s.d constraint. In practice, $L$ has a rectangular form ($L \in \mathbb{R}^{d \times r}$, $r \ll d$). However, OPML learns a square $d \times d$ matrix and obtains a closed-form solution with the $O(d^2)$ time complexity. Also, it adopts the Frobenius norm regularization term and the popular Hinge loss function. The interesting feature of OPML is the triplet sampling strategy which constructs the triplet from incoming data in an online manner.

OAHU[2] (Gao, Li et al. 2019) aims to dynamically adapt the complexity of the model and effectively utilizing the input constraints during the learning process. For this purpose, this method introduces the Adaptive-Bound Triplet Loss (ABTL) instead of commonly used Hinge loss. Also, it uses an over-complete neural network model and connects a different MEI (Metric Embedding Layer) to the input and each hidden layer of this network. The overall loss is considered as a weighted average loss of each MEI. Similar to previous approaches, it uses the Hedge algorithm to update the weights.

As seen, all studied online Distance/Similarity models assume that the given training information is perfect. However, this assumption may be wrong in practical machine vision applications where this information is collected from the Internet by crowdsourcing or similar techniques. Although some robust DML methods such as (Yang, Jin et al. 2010, Huang, Jin et al. 2012, Wang and Tan 2014, Wang, Nie et al. 2014, Wang and Tan 2018, Zabihzadeh, Monsefi et al. 2019, Rasheed, Zabihzadeh et al. 2020) are presented to address this emerging challenge. These methods are focused on batch-environments. Among these algorithms, only Bayesian approaches (Wang and Tan 2018, Zabihzadeh, Monsefi et al. 2019) can be extended for online settings. However, while Bayesian learning helps to avoid over-fitting in a small or a dataset with noisy features, it is less effective to deal with the more complicated problem i.e. label noise.

Many metric learning algorithms generate triplets from training data using the following *batch* procedure. Each data point $x_i$ is considered *similar* to its $k$ nearest neighbors with the same label (named *target neighbors* of $x_i$). Suppose $x_j$ is a *target neighbor* of $x_i$. The *imposter* of $x_i$ is any data point of a different class (i.e. $y_i \neq y_l$) which *violates* the following condition:

$$d(x_i, x_j) + margin < d(x_i, x_l)$$

where $d$ is a Distance/Similarity measure such as Euclidean. The data point $x_i$ is set dissimilar to any of its imposters. Then, the triplets are formed by the natural join of similar and dissimilar pairs. Figure 1 illustrates the concepts of *target neighbors* and *imposters*.

---

[1] One-Pass Metric Learning

[2] Online metric learning with Adaptive Hedge Update

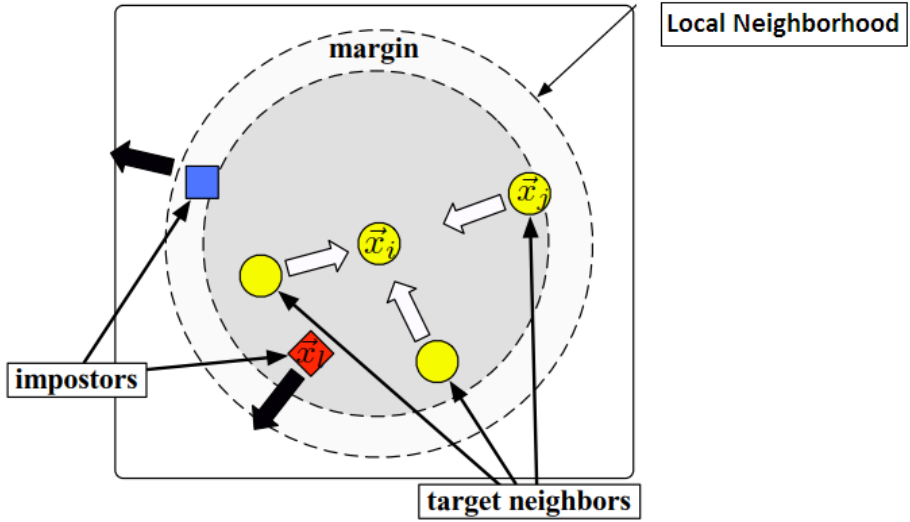

Figure 1: Illustration of target neighbors and imposters of the data point $x_i$ (Weinberger and Saul 2009)

Generating triplets using this procedure is both time and space consuming and is not feasible for online tasks. On the other hand, while online triplet construction adopted by OPML is very efficient in terms of computational cost, it does not consider the distribution and structure of data. Therefore, it has a lower performance in comparison with the batch algorithm.

## 3. Proposed Methods

As observed, many Distance/Similarity algorithms are based on the margin-based Hinge loss function ($l_{hinge}$). Let define the variable $z_t$ as follows:

$$z_t = \begin{cases} S_M(\boldsymbol{p_t}, \boldsymbol{p_t^+})^2 - S_M(\boldsymbol{p_t}, \boldsymbol{p_t^-})^2, & \text{For similarity} - \text{based methods} \quad (6) \\ d_M(\boldsymbol{p_t}, \boldsymbol{p_t^-})^2 - d_M(\boldsymbol{p_t}, \boldsymbol{p_t^+})^2, & \text{Mahalanobis} - \text{based methods} \quad (7) \end{cases}$$

The Hinge loss is then can be written as:

$$l(\boldsymbol{M}, (\boldsymbol{p_t}, \boldsymbol{p_t^+}, \boldsymbol{p_t^-})) = \max\{0, 1 - z_t\} \quad (8)$$

Figure 2 shows the loss function. As seen, the loss linearly grows for $z \leq 1$ with no bound. The unboundedness of the Hinge loss function causes the label noise and outlier data have a large effect in the training process leading to the poor performance of learned Distance/Similarity measure.

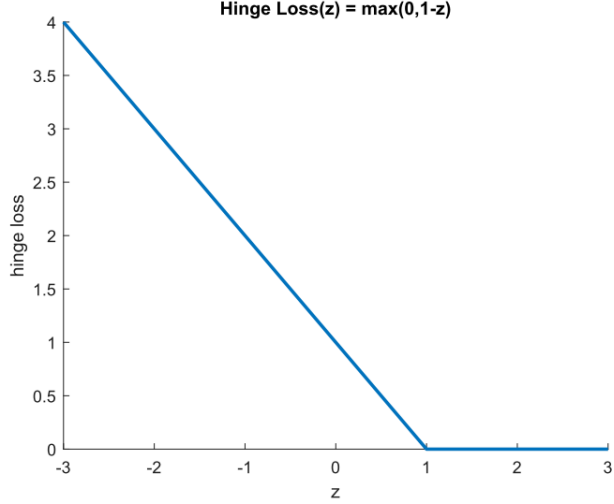

Figure 2: The margin-based Hinge loss function. The loss linearly grows for $z \leq 1$ with no bound.

Most existing Distance/Similarity learning methods can be formulated as follows:

$$M_{t+1} = \arg \min_{M} \text{reg}(M, M_t) + C l_{hinge}(z_t) \quad (9)$$
$$[\text{subject to } M \succcurlyeq 0]$$

Note that the constraint $M \succcurlyeq 0$ is not mandatory in all methods, so we enclose it by bracket. We can derive many existing methods from this generic optimization problem. For example, if we consider $\text{reg}(M, M_t) = \frac{1}{2}\|M - M_t\|_F^2$ and omit the $M \succcurlyeq 0$ constraint, then by defining $z_t$ according to (6), we obtain the OASIS (Chechik, Sharma et al. 2010) and OKS[1] (Xia, Hoi et al. 2014) optimization problems and if we consider $z_t$ equal to (7), the optimization problem (9) reduces to the OPML (Li, Gao et al. 2018). Similarly, if we set $\text{reg}(M, M_t) = \frac{1}{2}\|M - M_t\|_F^2$, enforce $M \succcurlyeq 0$, and define $z_t$ as (7), we reach to the optimization problem in (Wu, Hoi et al. 2016). Finally, if we set $\text{reg}(M, M_t) = D_{ld}(M, M_t) = \text{trace}(MM_t^{-1}) - \log\det(MM_0^{-1}) - d$ and drop the $M \succcurlyeq 0$ constraint, then we obtain the optimization problem of (Zhong, Zheng et al. 2017).

One approach to limit the effect of label noise data in PA-based problems (such as equation (9)) is to select a small value for the hyper-parameter C. However, it causes lower values for the adaptive learning rate in the PA-based algorithm. Instead, we propose to replace the Hinge loss function with the robust rescaled hinge loss defined as:

$$l_{rhinge}(z) = \beta \left[1 - \exp\left(-\eta l_{hinge}(z)\right)\right] \quad (10)$$

Figure 3 shows the diagram of the $l_{rhinge}(z)$ loss function with different values of $\eta$. In this function, $\eta$ is a rescaling parameter and $\beta = 1/(1 - \exp(-\eta))$ is just a normalizing constant that ensures $l_{rhinge}(0) = 1$. As seen, this loss function is more robust than the Hinge function against the outliers and data contaminated with label noise. We can adjust the degree of robustness using $\eta$ parameter. Also, the Hinge loss can be regarded as a special case of the rescaled Hinge. More specifically, $l_{rhinge}(z)$ becomes $l_{hinge}(z)$ as $\eta \to 0$.

---

[1] Online Kernel Similarity

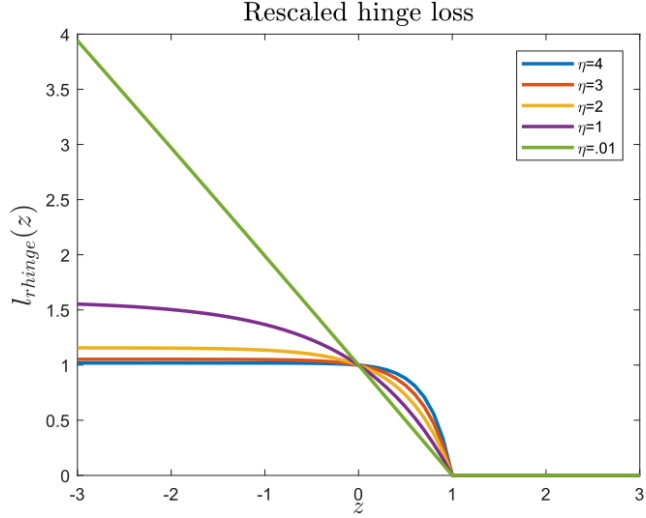

**Figure 3: The Robust Rescaled hinge loss function vs z with different $\eta$ values**

By replacing the Hinge loss function with the Rescaled Hinge loss in equation (9), we obtain the following optimization problem for online robust Distance/Similarity learning.

$$\boldsymbol{M_{t+1}} = \arg\min_{\boldsymbol{M}} \text{reg}(\boldsymbol{M}, \boldsymbol{M_t}) + C l_{rhinge}(z_t)$$

[subject to $\boldsymbol{M} \succcurlyeq 0$]

(11)

In the next subsection, we derive two efficient algorithms that efficiently solve the above optimization problem in online fashion.

### 3.1 The proposed robust methods

Since the rescaled hinge loss is not convex, we need an efficient algorithm to solve the optimization problem (11). The proposed algorithms are based HQ (Half Quadratic) which is an efficient alternating approach for optimizing the non-convex problem. The main idea of HQ is to add an auxiliary variable such as $v$ to the problem using Conjugate theory (Boyd, Boyd et al. 2004), such that the new optimization problem becomes quadratic respect to the main variable (with the same optimal solution as the original non-convex one).

Since $l_{rhinge}(z) = \beta \left[ 1 - \exp\left(-\eta l_{hinge}(z)\right) \right]$, we can obtain the following problem which is equivalent to (11).

$$\boldsymbol{M_{t+1}} = \arg\max_{\boldsymbol{M}} \; -\text{reg}(\boldsymbol{M}, \boldsymbol{M_t}) + C\beta \exp\left(-\eta l_{hinge}(z_t)\right)$$

[subject to $\boldsymbol{M} \succcurlyeq 0$]

(12)

According to the definition of conjugate function we have (refer to the Appendix A of (Xu, Cao et al. 2017)),

$$\exp\left(-\eta l_{hinge}(z)\right) = \sup_{v<0}\left(\eta l_{hinge}(z)v - g(v)\right) \tag{13}$$

where $g(v) = -v\log(-v) + v$, $(v < 0)$. By substituting equation (13) in (12), we obtain;

$$-\text{reg}(M, M_t) + C\beta \exp\left(-\eta l_{hinge}(z_t)\right)$$
$$= -\text{reg}(M, M_t) + C\beta \sup_{v_t<0}\left(\eta l_{hinge}(z_t)v_t - g(v_t)\right) \tag{14}$$
$$= \sup_{v_t<0}\left(-\text{reg}(M, M_t) + C\beta\left(\eta l_{hinge}(z_t)v_t - g(v_t)\right)\right)$$

The third relation in (14) holds since $-\text{reg}(M, M_t)$ is constant regarding $v$. Using (14), we can rewrite (12) as:

$$(M_{t+1}, v_t^*) = \arg\max_{M, v_t} \ -\text{reg}(M, M_t) + C\beta\left(\eta l_{hinge}(z_t)v_t - g(v_t)\right) \tag{15}$$
$$[\text{subject to } M \succcurlyeq 0]$$

To solve the above problem, we use the alternating optimization approach. First, given $M$, we optimize (13) over $v_t$ and then given $v_t$, we optimize it over $M$. Suppose $M^{(s)}$ is given (the superscript $s$ indicates the iteration number), then (15) is equivalent to:

$$v_t^{(s)} = \arg\max_{v_t} \ \left(\eta l_{hinge}(z_t)v_t - g(v_t)\right) \tag{16}$$

The above equation has a closed-form solution obtained by setting the derivative of it with respect to $v_t$ equal to zero.

$$v_t^{(s)} = -\exp\left(-\eta l_{hinge}(z_t)\right) \tag{17}$$

After obtaining $v_t^{(s)}$, we optimize the equation (15) respecting to $M^{(s+1)}$ as follows:

$$M = \arg\max_{M} \ -\text{reg}(M, M_t) + C\beta\eta v_t \, l_{hinge}(z_t) \tag{18}$$
$$[\text{subject to } M \succcurlyeq 0]$$

The above problem is equivalent to:

$$M = \arg\min_{M} \ \text{reg}(M, M_t) + C_t l_{hinge}(z_t) \tag{19}$$
$$\text{subject to } [M \succcurlyeq 0], \ l_{hinge}(z_t) \leq \xi, \quad \xi \geq 0$$

where $C_t = C\beta\eta(-v_t)$. The robustness of the optimization problem (19) can be explained using the penalty factor $C_t$. Suppose the current triplet $R_t$ contains label noise data, so the hinge function ($l_{hinge}(z_t)$) returns a big loss for $R_t$. Thus, $C_t = C\beta\eta(-v_t) = C\beta\eta \exp\left(-\eta l_{hinge}(z_t)\right)$ approaches to zero. Therefore, $R_t$ has a less effect in the learning process.

The obtained optimization problem, unlike existing models, assigns an adaptive weight ($C_t$) for each incoming triplet. By adjusting $\text{reg}(M, M_t)$, p.s.d constraint, and $z_t$, we can obtain a family of robust Distance/Similarity learning methods. For instance, we develop two proposed algorithms named Robust_OASIS and Robust_ODML[1]. These algorithms can be considered as robust versions of existing OASIS (Chechik, Sharma et al. 2010) and ODML (Wu, Hoi et al. 2016) respectively.

**Robust_OASIS**

This robust similarity-based algorithm can be derived from the general optimization problem (15) by the following settings:

$\text{reg}(M, M_t) = \frac{1}{2}\|M - M_t\|_F^2$, drop $M \succcurlyeq 0$ constraint, and define $z_t$ according to (6).

Then, the following optimization problem is achieved:

$$(M_{t+1}, v_t^*) = \arg\max_{M, v_t} -\frac{1}{2}\|M - M_t\|_F^2 + C\beta\left(\eta l_{hinge}(z_t)v_t - g(v_t)\right) \quad (20)$$

The solution of the above problem is obtained by iteratively computing $v_t$ from equation (17) and then optimizing $M$ by solving the following optimization problem.

$$M = \arg\min_{M} \frac{1}{2}\|M - M_t\|_F^2 + C_t\xi \quad (21)$$

subject to $l(p_t, p_t^+, p_t^-) = 1 - S_M(p_t, p_t^+) + S_M(p_t, p_t^-) \leq \xi, \quad \xi \geq 0$

The problem (21) has a similar solution to that obtained in (Chechik, Sharma et al. 2010).

$$M_{t+1} = M_t + \tau A_t \quad (22)$$

where $\tau = \min(C_t, \frac{l(p_t, p_t^+, p_t^-)}{\|A_t\|_F^2})$ and $A_t = p_t(p_t^+ - p_t^-)^T$

The main difference is that now the learning rate $\tau$ is bounded to the adaptive triplet weight $C_t$ instead of the fixed one ($C$) in the OASIS method. Algorithm 1 summarizes the steps of Robust-OASIS

---

[1] Robust Online Distance Metric Learning

**Algorithm 1.** Robust-OASIS

Inputs: $C, \eta, MaxHQIter$

Output: Similarity matrix $M$

    1. Initialize $M$ with the Identity matrix

    2. Set $v = 1$

    3. for $t = 1, 2, \ldots, T$

        3.1. Receive an input triplet $(p_t, p_t^+, p_t^-)$

        3.2. Compute $A_t = p_t(p_t^+ - p_t^-)^T$, and $\|A_t\|_F^2$

        3.3. for $s = 1, 2, \ldots, MaxHQIter$

            3.3.1. Update $v$ from (17)

            3.3.2 Compute the triplet weight $C_t = C\beta\eta(-v)$.

            3.3.3. Obtain the learning rate $\tau = \min(C_t, \frac{l(p_t, p_t^+, p_t^-)}{\|A_t\|_F^2})$

            3.3.4. Update $M$ from (22)

    end;

**Robust_ODML**

This robust Mahalanobis-based algorithm can be derived from the general optimization problem (15) by the following settings:

$\text{reg}(M, M_t) = \frac{1}{2}\|M - M_t\|_F^2$, enforce $M \succcurlyeq 0$ constraint, and define $z_t$ according to (6-2).

Then, we obtain the following optimization problem:

$$(M_{t+1}, v_t^*) = \arg\max_{M, v_t} -\frac{1}{2}\|M - M_t\|_F^2 + C\beta\left(\eta l_{hinge}(z_t)v_t - g(v_t)\right) \quad (23)$$

$$\text{subject to } M \succcurlyeq 0$$

Similar to the Robust-OASIS, we obtain the solution by iteratively computing $v_t$ from the equation (17) and then optimizing $M$ by solving the following optimization problem.

$$M = \arg\min_{M} \frac{1}{2}\|M - M_t\|_F^2 + C_t\xi \quad (24)$$

subject to $l(p_t, p_t^+, p_t^-) = 1 + d_M^2(p_t, p_t^+) - d_M^2(p_t, p_t^-) \leq \xi, \quad \xi \geq 0, \quad M \succcurlyeq 0$

The solution of the above problem is similar to that proposed in (Wu, Hoi et al. 2016).

$$M_{t+1} = M_t + \tau A_t \quad (25)$$

where $\tau = \min(C_t, \frac{l(p_t, p_t^+, p_t^-)}{\|A_t\|_F^2})$ and

$$A_t = (p_t - p_t^-)(p_t - p_t^-)^T - (p_t - p_t^+)(p_t - p_t^+)^T$$

Algorithm 2 summarizes the steps of Robust-ODML.

To enforce the p.s.d constraint, the simplest approach is to perform the full Eigen value decomposition of matrix $M$ and then set its negative Eigen values to zero. This approach requires $O(d^3)$ operations, so it is infeasible for high-dimensional DML tasks. Although some improved methods available in (Qian, Jin et al. 2015, Hamdan, Zabihzadeh et al. 2018, Rasheed, Zabihzadeh et al. 2020), we address this problem by developing the low-rank versions of the proposed algorithms in the next subsection.

---

**Algorithm2**. Robust-ODML

---

Inputs: $C, \eta, MaxHQIter$

Output: Distance matrix $M$

    1. Initialize $M$ with the Identity matrix

    2. Set $v = 1$

    3. for $t = 1,2,\ldots,T$

        3.1. Receive an input triplet $(p_t, p_t^+, p_t^-)$

        3.2. Compute $A_t = (p_t - p_t^-)(p_t - p_t^-)^T - (p_t - p_t^+)(p_t - p_t^+)^T$, and $\|A_t\|_F^2$

        3.3. for $s = 1,2,\ldots,MaxHQIter$

            3.3.1. Update $v$ from (17)

            3.3.2 Compute the triplet weight $C_t = C\beta\eta(-v)$.

            3.3.3. Obtain the learning rate $\tau = \min(C_t, \frac{l(p_t, p_t^+, p_t^-)}{\|A_t\|_F^2})$

            3.3.4. Update $M$ from (25)

    4. Enforce the p.s.d constraint: $M = psd(M)$

    end;

---

### 3.2 Low-rank Robust Distance/Similarity learning methods

The low-ranks versions of the proposed algorithms learn the projection matrix $L$ ($M = LL^T$) directly. Therefore, they automatically enforce p.s.d constraint. Also, in many practical applications, $L$ has a rectangular form ($L \in \mathbb{R}^{d \times r}$, $r \ll d$ is the rank of matrix $M$). Thus, the provided low-rank algorithms require fewer parameters. The optimization problem for low-rank online Distance/Similarity learning can be formulated as:

$$L_{t+1} = \arg\max_{L} \; - \text{reg}(L, L_t) + C\beta \left( \eta l_{hinge}(z_t) v_t - g(v_t) \right) \quad (26)$$

We can rewrite both the bilinear similarity and the Mahalanobis distance as functions of $L$ as follows:

$$S_L(p, q)^2 = p^T M q = p^T L L^T q = (L^T p)^T (L^T q) \quad (27)$$

$$d_L(p, q)^2 = (p - q)^T M (p - q) = (p - q)^T L L^T (p - q) = \|L^T p - L^T q\|_2^2 \quad (28)$$

Thus, the bilinear similarity learning is equivalent to finding a linear projection $L$ and then applying dot product to the inputs in the projected space. Similarly, Mahalanobis distance learning corresponds to compute the Euclidean distance after transforming the inputs by $L$.

The $z_t$ variable can be expressed in terms of $S_L$ and $d_L$ as:

$$z_t = \begin{cases} S_L(p_t, p_t^+)^2 - S_L(p_t, p_t^-)^2, & \text{For similarity} - \text{based methods} \quad (29) \\ d_L(p_t, p_t^-)^2 - d_L(p_t, p_t^+)^2, & \text{Mahalanobis} - \text{based methods} \quad (30) \end{cases}$$

Now, we can easily develop the proposed low-rank robust similarity learning algorithm named Robust-LOSL[1] from the generic optimization problem (26) with the following settings:

$$\text{reg}(L, L_t) = \frac{1}{2} \|L - L_t\|_2^2, \text{ define } z_t \text{ according to (29)}.$$

The obtained optimization problem can be solved by iteratively computing $v_t$ from the equation (17) and then optimizing $L$ by solving the following optimization problem:

$$M = \arg\min_{M} \; \frac{1}{2} \|L - L_t\|_F^2 + C_t \xi \quad (31)$$

$$\text{subject to } l(p_t, p_t^+, p_t^-) = 1 - S_L(p_t, p_t^+) + S_L(p_t, p_t^-) \leq \xi, \quad \xi \geq 0$$

The above optimization problem is non-convex. However, we can solve it efficiently by optimizing a simple linear neural network model depicted in Figure 4.

---

[1] Robust Low-rank Online Similarity Learning

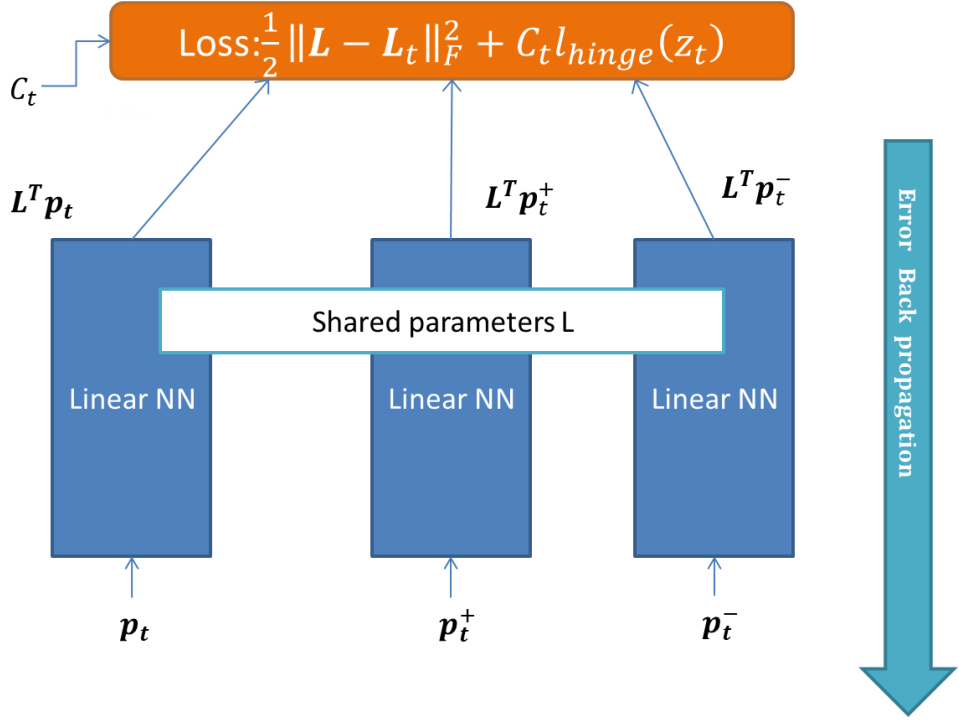

**Figure 4: The proposed neural network model for Robust Online Low-rank Distance/Similarity learning**

The sub-gradient of the loss function with respect to $L$ can be computed from the following equation:

$$\begin{aligned}
\frac{\partial l_t}{\partial L} &= (L - L_t) + C_t[\boldsymbol{p}_t \boldsymbol{p}_t^{-T} + \boldsymbol{p}_t^- \boldsymbol{p}_t^T - \boldsymbol{p}_t \boldsymbol{p}_t^{+T} - \boldsymbol{p}_t^+ \boldsymbol{p}_t^T]L \\
&= (L - L_t) - C_t[\boldsymbol{p}_t(\boldsymbol{p}_t^+ - \boldsymbol{p}_t^-)^T + (\boldsymbol{p}_t^+ - \boldsymbol{p}_t^-)\boldsymbol{p}_t^T]L \\
&= (L - L_t) - C_t[A_t + A_t^T]L \\
&\text{where } A_t = \boldsymbol{p}_t(\boldsymbol{p}_t^+ - \boldsymbol{p}_t^-)^T
\end{aligned} \qquad (32)$$

Thus, we can train the network from using backpropagation or more sophisticated algorithms such as Adams. The steps of Robust-LOSL are summarized in Algorithm3.

**Algorithm3.** Robust-LOSL

Inputs: $C, \eta, MaxHQIter$

Output: Similarity transformation matrix $\boldsymbol{L}$

    1. Initialize $\boldsymbol{L}$ with the Identity matrix

    2. Set $v = 1$

    3. for $t = 1,2, \dots, T$

        3.1. Receive an input triplet $(\boldsymbol{p}_t, \boldsymbol{p}_t^+, \boldsymbol{p}_t^-)$

        3.3. for $s = 1,2, \dots, MaxHQIter$

            3.3.1. Update $v$ from (16)

            3.3.2 Compute the triplet weight $C_t = C\beta\eta(-v)$.

            3.3.3. Optimize the network model parameterized by L using sub-gradient descent algorithm.

    end;

Similarly, we can derive the proposed low-rank robust distance learning algorithm named Robust-LODML[1] from the generic optimization problem (26) with the following settings:

$$\text{reg}(\boldsymbol{L}, \boldsymbol{L}_t) = \frac{1}{2}\|\boldsymbol{L} - \boldsymbol{L}_t\|_F^2, \text{ define } z_t \text{ according to (30)}.$$

We solve the obtained problem iteratively by computing $v_t$ from the equation (17) and then updating $\boldsymbol{L}$ by optimizing the neural network model presented in Figure 4. The sub-gradient of the loss function with respect to $\boldsymbol{L}$ can be computed from the following equation:

$$\begin{aligned}\frac{\partial l_t}{\partial \boldsymbol{L}} &= (\boldsymbol{L} - \boldsymbol{L}_t) + 2C_t[(\boldsymbol{p}_t - \boldsymbol{p}_t^+)(\boldsymbol{p}_t - \boldsymbol{p}_t^+)^T - (\boldsymbol{p}_t - \boldsymbol{p}_t^-)(\boldsymbol{p}_t - \boldsymbol{p}_t^-)^T]\boldsymbol{L} \\ &= (\boldsymbol{L} - \boldsymbol{L}_t) - 2C_t\boldsymbol{A}_t\boldsymbol{L} \\ &\text{where } \boldsymbol{A}_t = (\boldsymbol{p}_t - \boldsymbol{p}_t^-)(\boldsymbol{p}_t - \boldsymbol{p}_t^-)^T - (\boldsymbol{p}_t - \boldsymbol{p}_t^+)(\boldsymbol{p}_t - \boldsymbol{p}_t^+)^T\end{aligned} \quad (33)$$

Algorithm4 summarizes the steps of Robust_LODML. We can easily replace the linear module in the proposed low-rank model with a nonlinear deep neural network module. Thus, extending our methods for online deep Distance/Similarity learning is straightforward. Also, the experimental results in the next section confirm that the proposed low-rank methods reduce the computational cost significantly but also keeps the predictive performance of the learned measure.

---

[1] Robust Low-rank Online Distance Metric Learning

**Algorithm 4.** Robust-LODML

Inputs: $C, \eta, MaxHQIter$

Output: Distance transformation matrix $\boldsymbol{L}$

    1. Initialize $\boldsymbol{L}$ with the Identity matrix

    2. Set $v = 1$

    3. for $t = 1, 2, \ldots, T$

        3.1. Receive an input triplet $(\boldsymbol{p}_t, \boldsymbol{p}_t^+, \boldsymbol{p}_t^-)$

        3.3. for $s = 1, 2, \ldots, MaxHQIter$

            3.3.1. Update $v$ from (17)

            3.3.2 Compute the triplet weight $C_t = C\beta\eta(-v)$.

            3.3.3. Optimize the network model parameterized by L using sub-gradient descent algorithm.

    end;

### 3.3 Run Time Analysis

As seen, the proposed robust online Distance/Similarity learning model is general and can easily be applied to the existing online Distance/Similarity algorithms. Let $A$ be an online Distance/Similarity algorithm with the time complexity $T_A$. By applying our method to $A$, besides optimizing the Distance/Similarity measure, we require to compute the weight of the incoming triplet ($C_t$) using the equation (17). Although $C_t$ requires evaluation of $l_{hinge}(z_t)$, this evaluation is also needed for updating the metric. Therefore, it does not imply additional costs, and the overall time complexity of the robust method will be $O(MAXHQIter \times T_A)$. The experimental results confirm that the convergence of the alternating loop is fast, and the best results are obtained by taking $MAXHQIter \leq 3$ in all experiments. Therefore, the obtained robust method has the same time complexity as the corresponding algorithm ($A$).

### 3.4 Online Triplet Constructing Algorithm

Generating triplets using available batch algorithms is both time and space consuming. Also, the one-pass triplet constructing strategy adopted in OPML has low performance, especially in noisy environments. To this end, we propose an online triplet constructing algorithm named OCTG[1] which is not only very efficient but also effective in comparison with the available batch methods. By utilizing the distribution and clusters of input data, the proposed algorithm can effectively detect outliers and noisy label data. Therefore, its performance surpasses existing methods in noisy environments.

---

[1] Online Cluster-based Triplet Generator

Suppose $\{V_i | i = 1, 2, \ldots, K\}$ is the set of cluster centers initialized by a sample of data at the beginning of the online algorithm. In this paper, we use the k-means algorithm to obtain $c$ cluster centers per class in the dataset. OCTG receives incoming data $(x_t, y_t)$ at time step $t$ and finds its closest cluster center $V_t$ of the same class. Then, it considers any cluster center $V_i$ of a different class (i.e. $y_i \neq y_l$) which *violates* the following condition as an *imposter* (see Figure 5):

$$d(x_t, V_t) + margin < d(x_t, V_i)$$

The triplet set constructed at time step $t$ is formed as:

$$T_t = \{(x_t, V_t, V_i) | \text{ where } V_i \text{ is an imposter}\}$$

As seen, the proposed methods assign a weight to each incoming triplet. We assign the weight $w_t$ to $x_t$ equal to the minimum weights of the generated triplets. The small value for $w_t$ means that $x_t$ is a potential outlier or a noisy label instance. The weight and input data are then can be used to update the cluster centers using any existing online clustering methods.

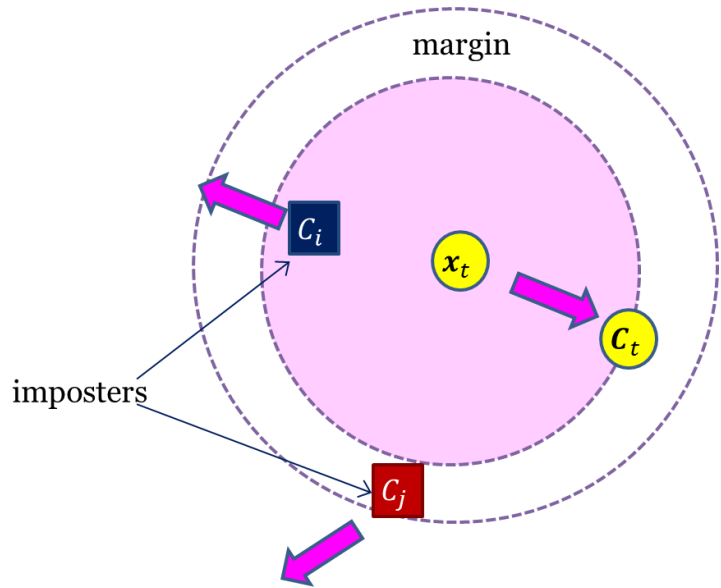

Figure 5- Illustration of imposters of the data point $x_t$

The obtained weights can be used to enhance the performance of any metric-based algorithms such as kNN or CBIR (Content-Based Information Retrieval) in noisy environments. For example, we use the following version of kNN named Robust-kNN (instead of standard kNN) to classify the objects in the experiments.

**Algorithm5.** Robust-kNN

Inputs:
- $X$: training examples
- $X\_test$: test instances
- $k$: k parameter in kNN
- $w$: instance weights vector
- $\eta$: Percentage of noisy labels

1. Sort data according to $w$ values
2. Remove $\eta$ percent of data in **X** with the lowest weights
3. **for each** instance $x\_test$ in $X\_test$
    3.1 Compute $k$ nearest neighbors in Set $X$
    3.2 Predict label of $x\_test$ based on the weighted majority voting of its neighbors.

Figure 6 depicts the system flow of the proposed learning/test schemes.

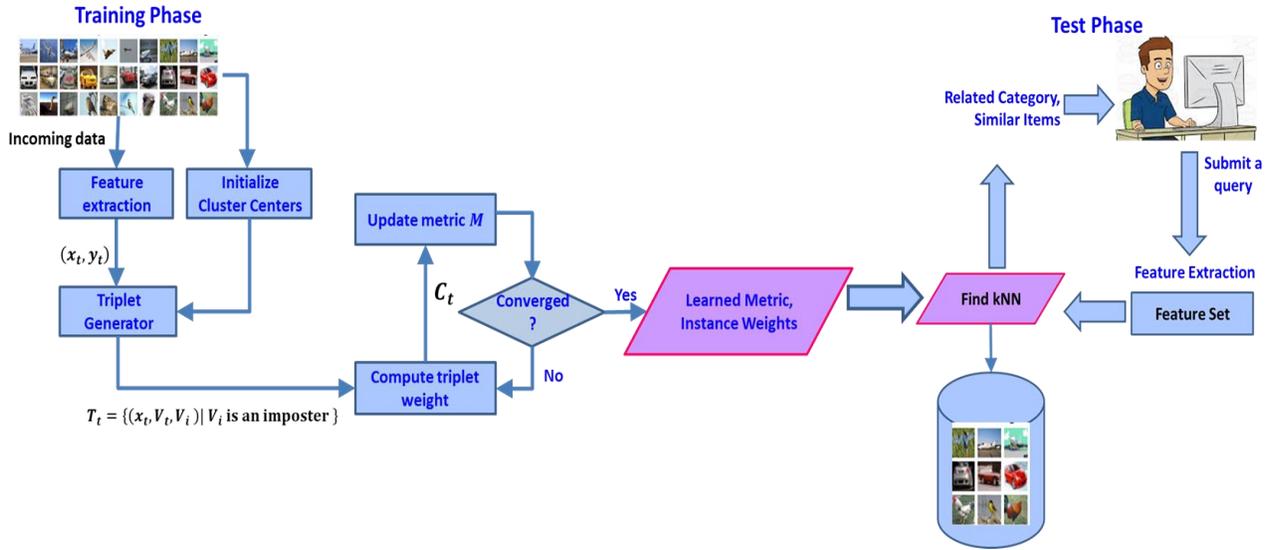

**Figure 6: The system flow of the proposed learning/test schemes**

## 4. Experimental Results

This section deals with the experiments performed to evaluate the performance of proposed methods in noisy environments. We first study the effect of label noise on the generated triplets and then discuss how these noisy triplets affect the performance of online Distance/Similarity methods. Afterward, we evaluate the performance of proposed methods on real datasets at different levels of label noise. The results are compared with peer Distance/Similarity methods.

### 4.1 Effect of Label Noise on the Generated Triplets

As depicted in Figure 7, we can distinguish between three different types of noisy triplets: *anchor*, *positive*, and *negative noisy triplet*.

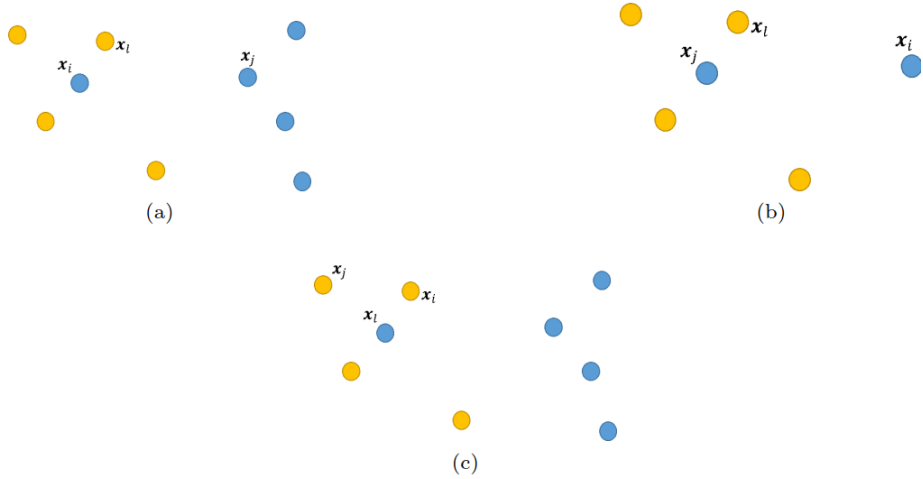

Figure 7: Three different types of noisy triplets in the form $(x_i, x_j, x_l)$: (a) Anchor noisy triplet where $x_i$ is contaminated with label noise, (b) Positive noisy triplet where $x_j$ has label noise, and (C) Negative noisy triplet where $x_l$ has wrong label.

To study the effects of different types of noisy triplets, we apply 10% label noise to the Wine dataset. The noisy dataset is visualized using the T-SNE algorithm (Maaten and Hinton 2008) in Figure 8.

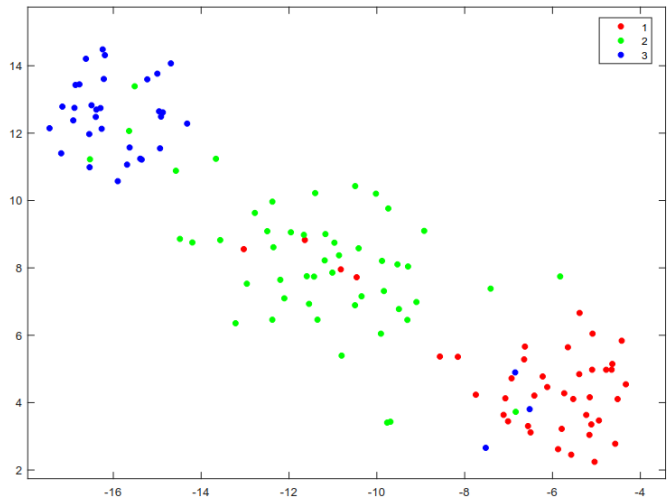

Figure 8: T-SNE Visualization of the Wine dataset after applying 10% label noise

The statistics of the generated triplets using both batch and OCTG methods are summarized in Table 1.

Table 1-Statistics of generated triplets in the Wine dataset contaminated with 10% label noise

| Method<br>Feature | Batch | | OCTG (Ours) | |
|---|---|---|---|---|
| | # | Mean Hinge loss | # | Mean Hinge loss |
| Instances | 178 | - | 178 | - |
| Classes | 3 | - | 3 | - |
| Triplets | 413 | 0.92 | 140 | 0.71 |
| Normal triplets | 131 | 0.85 | 105 | 0.39 |
| Noisy triplets | 282 | 0.96 | 35 | 1.67 |
| Anchor noisy triplets | 38 | 1.01 | 35 | 1.67 |
| Positive noisy triplets | 23 | 1.02 | 0 | - |
| Negative noisy triplets | 249 | 0.95 | 0 | - |

As the results in Table 1 indicate, by applying only 10% label noise, 68% of generated triplets by the batch method are contaminated. On the other hand, OCTG only constructs 25% contaminated triplets (just from anchor noisy type). The generated noisy triplets by OCTG have large losses in comparison with the normal ones (1.67 vs 0.39). Hence, the proposed robust methods assign very small weights $\left(C_t = C\beta\eta \exp\left(-\eta l_{hinge}(z_t)\right)\right)$ to them in the learning process. That causes they have less effect on the learned metric.

To analyze the effect of different types of triplet noise in a typical DML task, we run the ODML (Wu, Hoi et al. 2016) with the following settings on the generated triplets by the batch method.

**ODML**: The ODML algorithm

**Ideal ODML**: The ideal algorithm which knows the noisy triplets in advance and so ignores them in the training process

**Anchor Ideal ODML:** The ideal algorithm which knows only the anchor noisy triplets in advance

**Pos Ideal ODML:** The ideal algorithm which knows only the positive noisy triplets in advance

**Neg Ideal ODML:** The ideal algorithm which knows only the negative noisy triplets in advance

In this experiment, we divide the dataset into train/test with 70/30 ratio and run the above algorithms 10 times on the dataset. Figure 9 depicts the mean of obtained results by various algorithms.

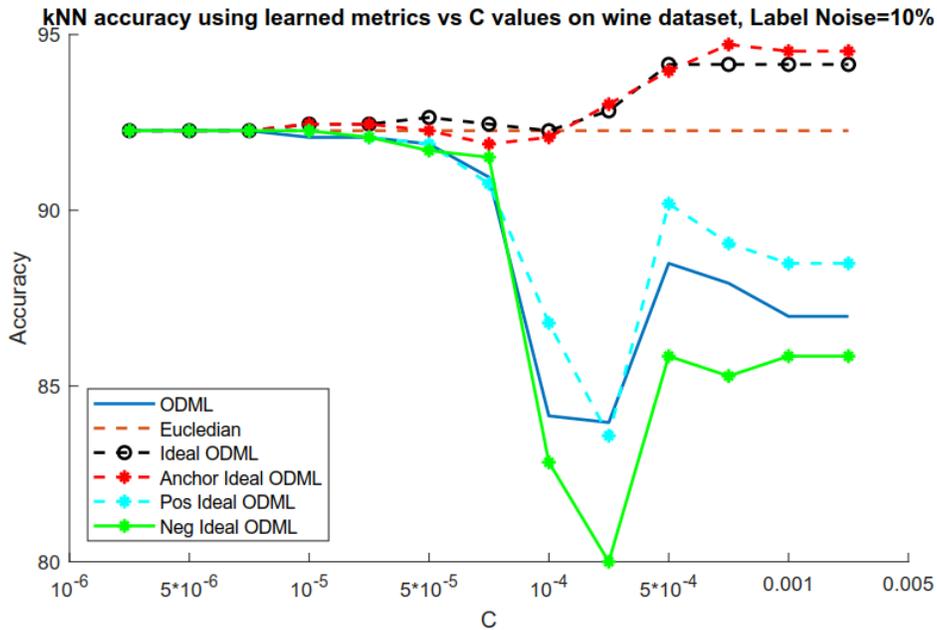

**Figure 9:** The kNN accuracy of the learned metric of various algorithms in the Wine dataset with 10% label noise

For small values of C, the results indicate that the learned metric by ODML has no meaningful difference with that of Euclidean. For large values of C, ODML performs worse than Euclidean and its accuracy substantially degrades in the noisy environment. Also, among the ideal methods (cannot be implemented in practice), the *Anchor Ideal ODML* has the same performance as Ideal ODML and others (*Pos Ideal ODML*, *Neg Ideal ODML*) are ineffective. Thus, the main reason for low performance in this DML task is anchor noisy triplets.

We repeat the experiment by running Robust-LODML using the triplets generated by our mechanism. The results are depicted in Figure 10 and Figure 11. As the results show, the proposed method is robust against label noise and its performance surpasses Euclidean metric even for the large values of C. Also, Robust-LODML effectively identified the contaminated instances and considerably reduces their weights in the training process.

The results are obtained by using only one dataset. In the next subsections, we evaluate the proposed methods on the variety of datasets at different levels of label noise. Also, the results are compared with state-of-the-art methods.

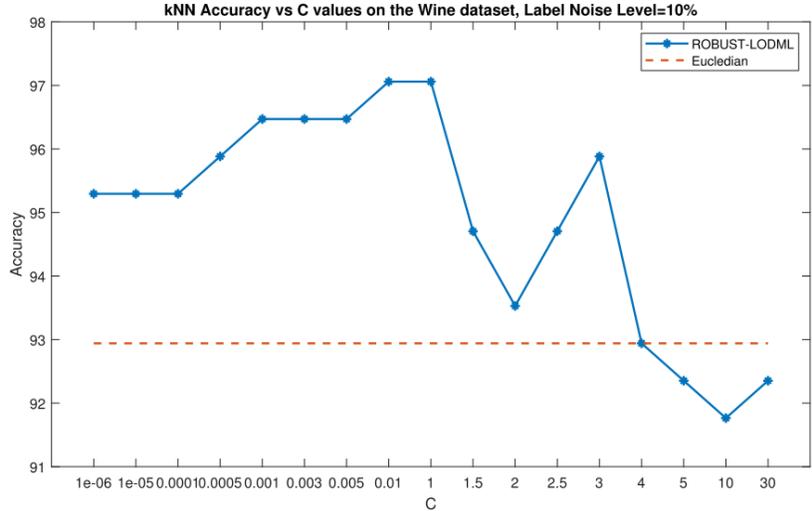

**Figure 10: The kNN accuracy of the learned metric by Robust-LODML algorithm ($\eta = 3$) in the Wine dataset with 10% label noise**

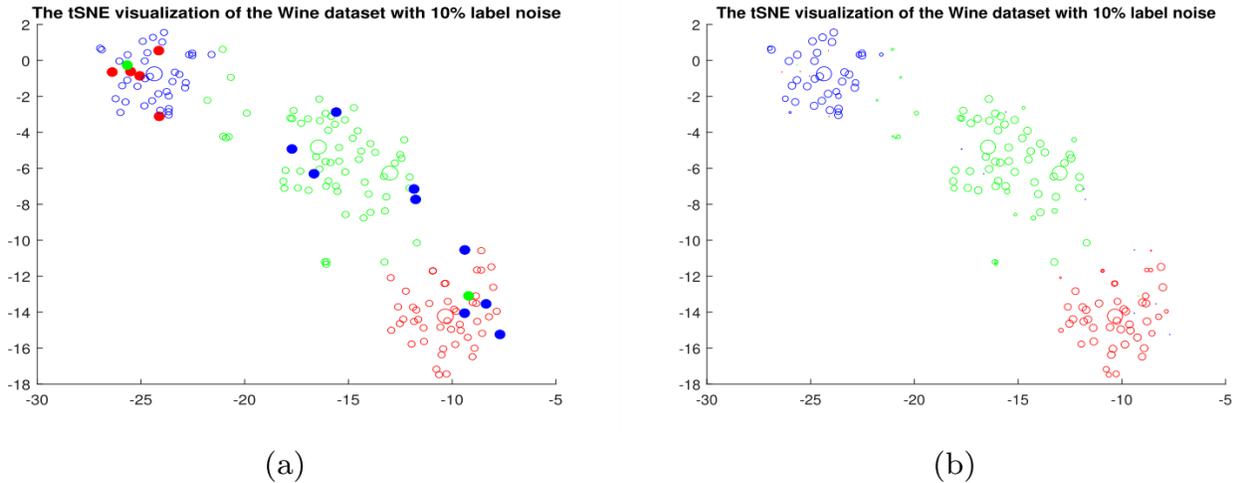

**Figure 11: The tSNE visualization of the Wine dataset with 10% label noise where data points are displayed (a) with equal size (b) with the size proportional to the weight**

### 4.2 Experimental Setup

Table 2 summarizes the statistics of evaluated datasets in the experiments. The data in all datasets except *Letters* is normalized so that the mean and standard deviation of each attribute becomes 0 and 1, respectively. Also, the dimension of images in *Extended Yale Faces* is reduced to 100 by applying PCA to alleviate the noise effects. The parameter *d* in Table 2 denotes the input dimension after dimension reduction.

In the experiments, triplet side information is generated using OCTG for the proposed methods while the one-pass triplet construction in (Li, Gao et al. 2018) is adopted for the other methods.

**Table 2-Statistics and explanations of evaluated datasets**

| Data Set | #classes | n | #dim | d | Description |
|---|---|---|---|---|---|
| Wine (Lichman 2013) | 3 | 178 | 13 | 13 | Standard UCI classification dataset. https://archive.ics.uci.edu/ml/datasets/wine |
| Letters (Lichman 2013) | 26 | 20,000 | 16 | 16 | includes 20,000 examples of 26 English capital letters. Images of letters are generated from 20 different fonts and then 16 numerical attributes are extracted from these images. https://archive.ics.uci.edu/ml/datasets/letter+recognition |
| Extended Yale Faces (Lee, Ho et al. 2005) | 38 | 2,414 | 1,024 | 200 | is a standard face recognition dataset contains 2,414 face images of 38 classes. For each person, at most 64 images are taken under extreme illumination conditions. http://vision.ucsd.edu/~iskwak/ExtYaleDatabase/ExtYaleB.html |
| Ionosphere (Lichman 2013) | 10 | 351 | 34 | 33 | Standard UCI classification dataset. https://archive.ics.uci.edu/ml/datasets/Ionosphere |
| WDBC (Lichman 2013) | 2 | 569 | 32 | 30 | Breast Cancer Wisconsin (Diagnostic) Data Set https://archive.ics.uci.edu/ml/datasets/Breast+Cancer+Wisconsin+(Diagnostic) |
| Australian | 2 | 690 | 14 | 14 | used in a competition on click-through rate prediction jointly hosted by Avazu and Kaggle in 2014. The participants were asked to from the first 10 days of advertising log, estimate the click probability for the impressions on the 11th day. https://www.csie.ntu.edu.tw/~cjlin/libsvmtools/datasets/binary.html |
| German Credit | 2 | 1000 | 24 | 24 | Each instance represents a person who takes a credit by a bank and is classified as good or bad credit risks according to the set of attributes. https://www.kaggle.com/uciml/german-credit |

The results are obtained by k-fold cross validation (k=5 is set for Letters and Extended Yale Faces and k=10 for other datasets). The results are compared with peer distance-based methods: ODML (Wu, Hoi et al. 2016), LPA-ODML[1](Hamdan, Zabihzadeh et al. 2018), and OPML (Li, Gao et al. 2018).

The hyperparameters of the competing methods are adjusted by k-fold cross-validation as follows. The parameter $C$ in ODML and the proposed methods are selected from $(10^{-6}, 30)$. The $\eta$ in the proposed methods is chosen from the range $(0.01, 5)$. Also, $\lambda$ in OPML is selected from $(10^{-6}, 0.05)$. To evaluate the performance of the learned metrics, we adopt the kNN classifier with $k = 3$ in the experiments.

### 4.3 Results and Analysis

Table 3 presents the classification accuracy of the kNN using the learned metrics of the competing methods. Here, the parameter $nl$ shows label noise level (in percent). The results are obtained by k-fold cross-validation on these datasets. Also, Figure 7 depicts the mean of k-fold cross validation accuracy of competing methods versus $nl$ (ranging from 0% to 20%).

---

[1] Local Passive/Aggressive Online Distance Metric Learning

Table 3- The classification rate of the kNN classifier using the learned metric of the competing methods

| Data Set | nl % | Robust-LODML | Robust-ODML | LPA-ODML | ODML | OPML | Euclidean |
|---|---|---|---|---|---|---|---|
| Wine | 0 | **97.65+-4.11** | **96.47+-6.32** | **97.06+-4.16** | 98.29+-1.56 | 97.06+-4.16 | 95.29+-5.41 |
|  | 5 | **97.65+-3.04** | **97.06+-3.10** | **97.06+-5.00** | 96.00+-4.78 | 95.88+-4.84 | 93.53+-5.154 |
|  | 10 | **96.47+-4.96** | **97.06+-5.00** | **96.47+-4.11** | 93.14+-3.26 | 96.47+-4.11 | 94.12+-4.80 |
|  | 15 | **97.65+-4.11** | **95.88+-5.58** | **94.71+-6.47** | 90.86+-6.52 | 94.12+-5.55 | 92.35+-6.82 |
|  | 20 | **95.29+-6.68** | **95.29+-5.41** | **90.00+-6.82** | 89.14+-3.73 | 91.18+-8.43 | 85.88+-8.41 |
| Letters | 0 | 96.80+-0.25 | 96.68+-0.37 | 96.88+-0.25 | 96.76+-0.29 | 96.78+-0.34 | 95.39±0.36 |
|  | 5 | 96.41+-0.35 | 95.85+-0.61 | 96.08+-0.29 | 95.98+-0.28 | 96.02+-0.37 | 94.53±.50 |
|  | 10 | 95.39+-0.35 | 94.36+-0.44 | 93.57+-0.34 | 94.08±0.32 | 94.29+-0.31 | 92.64±0.51 |
|  | 15 | 94.19+-0.40 | **93.27+-0.63** | 91.78+-0.28 | 91.53+-0.71 | 91.57+-0.40 | 90.03+-0.55 |
|  | 20 | 93.18+-0.48 | **92.20+-1.04** | 88.69+-0.30 | 88.46+-0.40 | 88.32+-0.32 | 86.67+-0.81 |
| Extended Yale Faces | 0 | **96.02+-0.31** | **95.52+-1.12** | **93.94+-0.90** | **93.82+-.82** | 93.57+-0.88 | 93.36+-0.89 |
|  | 5 | **95.56+-0.45** | **94.27+-1.12** | **92.86+-0.84** | **92.82+-1.05** | 92.53+-0.62 | 92.57+-0.27 |
|  | 10 | **94.94+-1.01** | **93.69+-1.02** | **92.78+-1.24** | **91.70+-1.41** | 90.95+-1.23 | 91.54+-0.86 |
|  | 15 | **94.90+-1.20** | **92.70+-1.78** | **91.33+-0.90** | **88.51+-1.19** | 88.71+-1.32 | 88.63+-1.03 |
|  | 20 | **93.11+-1.40** | **92.37+-0.68** | **88.88+-1.26** | **85.56+-0.91** | 85.23+-1.28 | 85.56+-0.91 |
| Ionosphere | 0 | **93.14+-3.35** | **92.00+-3.24** | **94.00+-4.94** | 90.29+-4.09 | 86.57+-5.05 | 84.86+-3.29 |
|  | 5 | **93.14+-4.09** | **91.43+-5.39** | **91.71+-4.75** | 89.71+-3.61 | 87.43+-4.30 | 86.00+-3.41 |
|  | 10 | **90.86+-4.82** | **91.43+-4.47** | **88.86+-4.56** | 87.71+-3.82 | 86.57+-3.31 | 84.57+-4.89 |
|  | 15 | **90.00+-4.90** | **89.14+-9.31** | **87.14+-6.06** | 87.71+-5.23 | 84.35+-7.31 | 81.43+-2.26 |
|  | 20 | **88.86+-4.75** | **88.00+-5.35** | **83.71+-4.68** | 84.29+-6.35 | 82.32+-7.25 | 79.43+-6.29 |
| WDBC | 0 | **95.71+-2.41** | **95.36+-2.55** | **95.18+-3.04** | 95.00+-2.64 | 95.00+-3.84 | 92.86+-3.15 |
|  | 5 | **94.29+-3.01** | **95.54+-2.70** | **94.82+-2.59** | 93.93+-3.17 | 93.75+-3.88 | 92.32+-3.37 |
|  | 10 | **94.46+-3.31** | **93.57+-2.41** | **93.04+-2.59** | 92.68+-2.97 | 93.57+-2.94 | 89.11+-3.81 |
|  | 15 | **94.11+-3.04** | **92.32+-3.67** | **90.18+-3.69** | 88.39+-6.08 | 88.93+-2.35 | 85.71+-5.26 |
|  | 20 | **91.07+-2.92** | **92.50+-4.67** | **85.89+-3.62** | 85.00+-5.60 | 85.89+-3.81 | 83.93+-6.63 |
| Australian | 0 | **85.51+-5.11** | **86.23+-3.82** | **86.67+-4.72** | **85.51+-3.98** | 83.62+-6.11 | 82.03+-5.43 |
|  | 5 | **86.09+-5.17** | **87.25+-5.24** | **85.94+-5.68** | 84.93+-4.44 | 83.33+-5.08 | 81.30+-5.31 |
|  | 10 | **84.78+-2.49** | **86.67+-4.72** | **83.77+-5.01** | 81.45+-4.62 | 81.45+-4.92 | 78.99+-7.43 |
|  | 15 | **85.07+-4.88** | **85.07+-4.32** | **82.32+-4.92** | 78.26+-5.68 | 81.45+-3.54 | 76.81+-5.76 |
|  | 20 | **85.51+-3.92** | **85.80+-3.19** | **79.42+-5.10** | 74.64+-6.84 | 78.84+-4.94 | 73.19+-5.12 |
| German Credit | 0 | **74.60+-2.07** | **74.70+-2.21** | **76.10+-3.60** | **73.90+-2.38** | 72.20+-3.79 | 69.40+-4.14 |
|  | 5 | **74.60+-2.55** | **73.40+-5.42** | **73.50+-4.40** | 72.40+-4.20 | 70.30+-4.50 | 67.90+-5.26 |
|  | 10 | **73.20+-4.44** | **74.00+-3.40** | **72.70+-3.13** | 71.10+-5.22 | 69.40+-5.21 | 66.50+-5.99 |
|  | 15 | **73.20+-3.58** | **73.30+-4.90** | **71.20+-5.09** | 68.60+-2.95 | 65.20+-3.88 | 64.30+-4.79 |
|  | 20 | **72.60+-3.50** | **71.50+-4.14** | **70.30+-5.54** | 65.70+-3.68 | 63.30+-5.14 | 62.30+-6.20 |

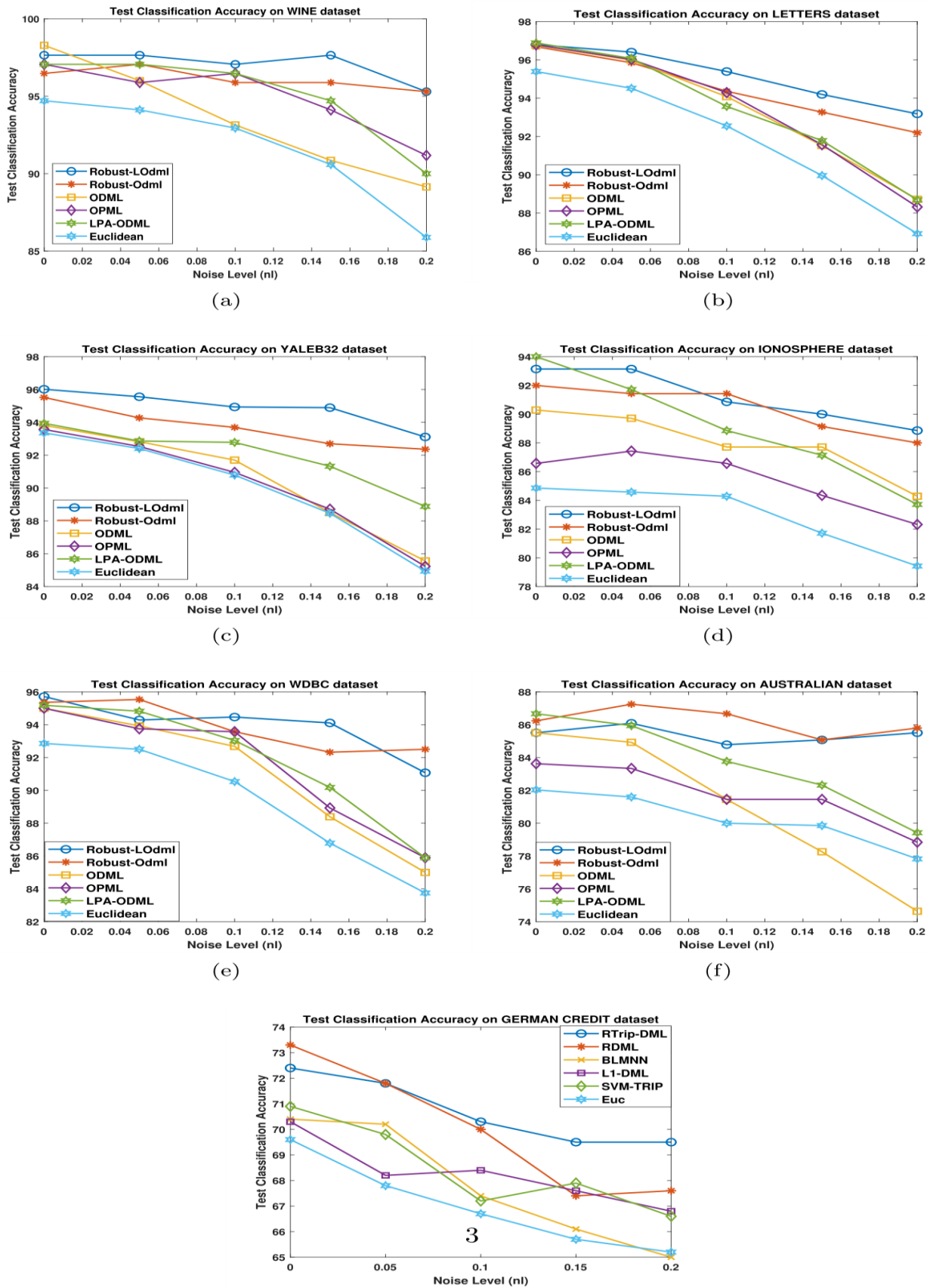

**Figure 12:** Comparison of the classification accuracy of RDML with other DML methods versus label noise.

As the results in Table 3 and Figure 12 indicate, the proposed robust methods (i.e. Robust-ODML and Robust-LODML) significantly outperform other DML methods in the presence of label noise. Also, the performance of these methods declines slowly than other ones with the increase of noise level. That confirms our claim that using robust loss function jointly with the proposed robust sampling preserves the discrimination of learned metric in a noisy environment. In addition, the low-rank version of the proposed method (i.e. Robust-LODML) almost has the same accuracy as Robust-ODML while reduces the computational cost significantly. In the next subsection, we evaluate our proposed methods in a more challenging dataset used to detect COVID-19 patients from Chest-X-ray images.

**Dataset Description**

The dataset used in our experiments is publicly available in the kaggle repository[1] (Chowdhury, Rahman et al. 2020). Figure 13 depicts some examples from both classes. It contains 219 COVID-19 cases and 1341 normal images. As seen, the dataset is imbalanced and is too small to train a deep CNN model from scratch.

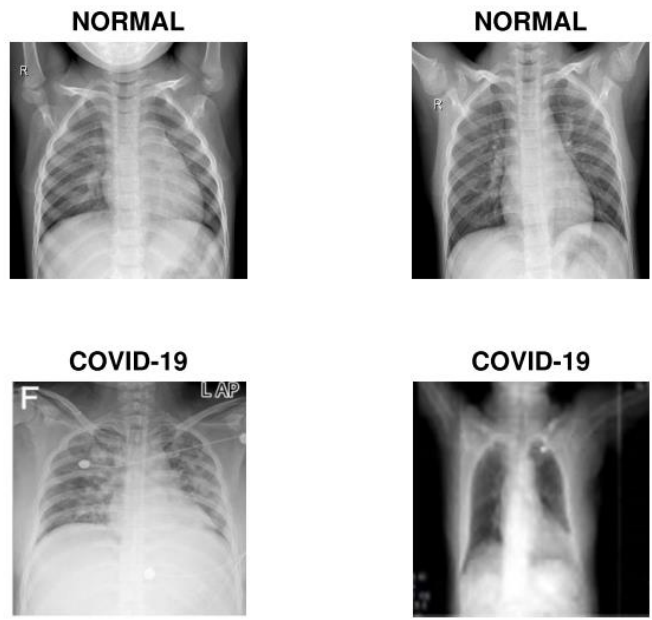

Figure 13: Four images from the COVID-19 dataset. First row: Normal cases, Second row: COVID-19 patients

**Experimental Setup**

To extract features from the images, we use the pretrained Resnet18 (He, Zhang et al. 2016). This network is trained on the ImageNet dataset (with 1.4 million labeled images and 1,000 different classes). It has 71 layers and the input layer requires input images of size 224-by-224-

---

[1] https://www.kaggle.com/tawsifurrahman/covid19-radiography-database?select=COVID-19+Radiography+Database

by-3. We resize the images to the specified size and obtain 512 features from the global pooling layer, *'pool5',* at the end of the model.

We use 5-fold cross-validation to obtain the results in the experiments. The main concern in this task is to limit the number of missed COVID-19 cases. Hence, in addition to accuracy, we utilize a variety of metrics to evaluate our work. These metrics are Sensitivity (Recall), Precision, F1 Score, and G-mean (Geometric-mean). Here, COVID-19 and Normal are considered as positive and negative, respectively. The metrics are defined as follows:

$$Accuracy = (TP + TN) / All\ Predictions \tag{34}$$

$$Sensitivity\ (Recall) = TP / (FN + TP) \tag{35}$$

$$Precision = TP / (TP + FP) \tag{36}$$

$$F1 - Score = 2\ (Precision \times Sensitivity) / (Precision + Sensitivity) \tag{37}$$

$$Specificity = TN / (TN + FP) \tag{38}$$

$$G - mean = \sqrt{Sensitivity \times Specificity} \tag{39}$$

## Results and Analysis

Table 4 presents the classification results of the kNN using the learned metrics in the different levels of label noise. The results of both *sensitivity* and *precision* of the competing methods versus noise level are shown in Figure 13 (a). Since sensitivity is more important in this task, we multiply it by 2. Also, Figure 13 (b) presents the G-mean results versus noise level. The high value of G-mean indicates that accuracy in both classes is high and balanced.

As the results indicate, the proposed methods achieve high sensitivity for COVID-19 patients in noisy environments. It is very important since the primary goal of this task is to limit the number of misclassified COVID-19 cases as much as possible. For example, the Confusion matrices of the proposed methods at *Noise Level* = 20% are shown in Table 5. As seen, only 1.8 and 1 (as the average of 5-fold cross validation) COVID-19 patients are misclassified as Normal by the proposed methods. Also, our methods obtain good precision (or predictive positive value). High precision is important since high FP (False Positive) increases the burden of the healthcare system for additional care and tests such as PCR (Polymerase Chain Reaction). Therefore, based on the results we can conclude the proposed methods perform well in detecting COVID-19 cases in the presence of label noise. However, the difference between *sensitivity* and *specificity* values indicate further improvements are possible by adopting *balancing techniques* in this imbalanced dataset.

Table 4- Classification metrics of kNN using the learned metrics of competing methods on COVID-19 dataset

| Method | nl % | Accuracy | Sensitivity | Precision | Specificity | G-mean | F1-Score |
|---|---|---|---|---|---|---|---|
| Robust-ODML | | **99.23+-0.66** | **95.35+-3.52** | **99.65+-0.78** | 99.92+-0.18 | 97.59+-1.83 | 97.43+-1.99 |
| Robust-LODML | | **99.42+-0.57** | **96.54+-3.80** | **99.50+-1.12** | 99.93+-0.17 | 98.20+-1.96 | 97.97+-2.16 |
| LPA-ODML | 0 | **99.49+-0.37** | **97.33+-3.06** | **99.13+-1.25** | 99.85+-0.21 | 98.57+-1.50 | 98.19+-1.32 |
| ODML | | **99.36+-0.23** | **97.33+-3.06** | **98.10+-2.18** | 99.70+-0.31 | 98.50+-1.42 | 97.66+-0.88 |
| OPML | | **99.29+-0.42** | **96.62+-2.61** | **98.60+-2.29** | 99.78+-0.33 | 98.18+-1.25 | 97.56+-1.10 |
| Robust-ODML | | 99.10+-0.48 | 96.11+-2.61 | 97.75+-1.30 | 99.62+-0.28 | 97.84+-1.30 | 96.90+-1.13 |
| Robust-LODML | | 98.97+-0.35 | 96.99+-3.01 | 95.83+-1.25 | 99.33+-0.17 | 98.14+-1.47 | 96.37+-1.08 |
| LPA-ODML | 5 | 99.17+-0.62 | 94.71+-4.12 | 99.52+-1.06 | 99.93+-0.17 | 97.26+-2.13 | 97.02+-2.32 |
| ODML | | 98.97+-0.42 | 96.50+-3.37 | 96.29+-2.26 | 99.40+-0.33 | 97.93+-1.64 | 96.34+-1.39 |
| OPML | | 98.53+-0.49 | 93.81+-3.21 | 95.72+-5.45 | 99.34+-0.79 | 96.52+-1.41 | 94.62+-1.97 |
| Robust-ODML | | **98.46+-0.80** | **94.90+-2.03** | **94.78+-3.34** | 99.00+-1.01 | 96.92+-0.87 | 94.79+-1.32 |
| Robust-LODML | | **98.46+-0.83** | **94.97+-2.98** | **94.31+-3.53** | 99.02+-0.64 | 96.96+-1.66 | 94.62+-2.78 |
| LPA-ODML | 10 | **98.21+-1.23** | **88.95+-7.19** | **99.07+-1.30** | 99.85+-0.21 | 94.18+-3.73 | 93.59+-3.62 |
| ODML | | **97.50+-0.89** | **92.53+-4.39** | **90.20+-7.81** | 98.37+-1.22 | 95.38+-2.00 | 91.09+-3.33 |
| OPML | | **98.08+-0.60** | **90.53+-5.19** | **96.04+-4.68** | 99.41+-0.66 | 94.83+-2.44 | 93.00+-1.19 |
| Robust-ODML | | **97.95+-0.66** | **96.96+-2.46** | **89.16+-4.22** | 98.14+-0.62 | 97.54+-1.32 | 92.84+-2.59 |
| Robust-LODML | | **97.76+-1.96** | **95.68+-3.36** | **90.52+-7.72** | 98.09+-1.80 | 96.87+-2.46 | 92.97+-5.58 |
| LPA-ODML | 15 | **98.21+-0.98** | **90.78+-3.68** | **96.87+-2.13** | 99.47+-0.45 | 95.01+-1.99 | 93.70+-2.35 |
| ODML | | **95.90+-0.83** | **93.69+-4.61** | **79.84+-5.70** | 96.29+-0.75 | 94.96+-2.35 | 86.09+-3.79 |
| OPML | | **96.03+-0.18** | **93.41+-2.71** | **80.84+-4.63** | 96.50+-0.48 | 94.93+-1.16 | 86.56+-1.86 |
| Robust-ODML | | **96.73+-0.69** | **97.94+-2.03** | **82.06+-5.47** | 96.57+-0.93 | 97.24+-0.82 | 89.17+-2.67 |
| Robust-LODML | | **97.31+-1.03** | **96.08+-4.67** | **86.35+-5.73** | 97.54+-0.97 | 96.78+-2.38 | 90.82+-3.51 |
| LPA-ODML | 20 | **97.50+-1.66** | **89.82+-4.86** | **92.88+-7.40** | 98.80+-1.30 | 94.18+-2.90 | 91.23+-5.36 |
| ODML | | **92.63+-0.93** | **90.45+-3.36** | **67.32+-5.37** | 92.99+-0.71 | 91.70+-1.91 | 77.09+-3.93 |
| OPML | | **92.44+-1.41** | **91.40+-5.26** | **66.55+-7.70** | 92.64+-1.62 | 91.98+-2.59 | 76.75+-5.41 |

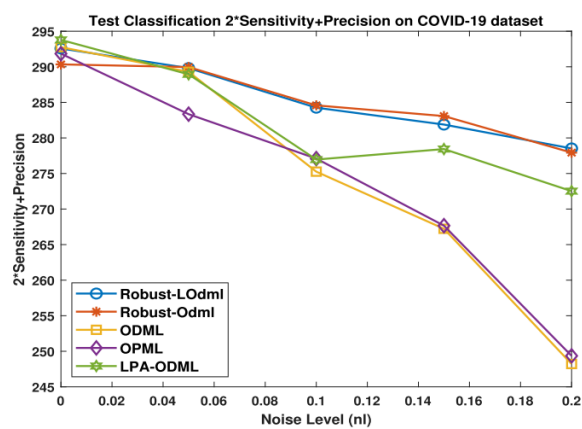

(a)

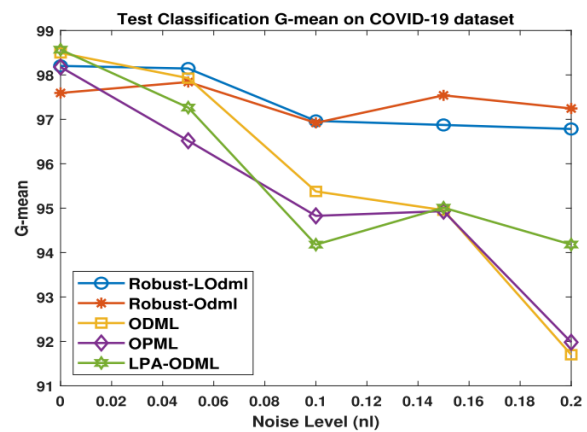

(b)

Figure 14- 2×Sensitivity+Precision and G-means of the competing methods on COVID-19 dataset.

**Table 5- Mean of confusion matrices of proposed methods obtained by 5-fold cross validation on COVID-19 dataset with label noise=20%**

|  |  | Predicted Positive (COVID-19) | Predicted Negative (Normal) |
|---|---|---|---|
| **Robust-LODML** | Actual Positive (COVID-19) | 42.00 | 1.8 |
|  | Actual Negative (Normal) | 6.60 | 261.60 |
| **Robust-ODML** | Actual Positive (COVID-19) | 42.8 | 1.00 |
|  | Actual Negative (Normal) | 9.2 | 259.00 |

## 5. Conclusion and Future Work

Existing online Distance/Similarity learning methods usually formulated by the Hinge loss and so are not robust against outliers and label noise data. Also, they often have the wrong assumption that training triplets or pairwise constraints exist in advance. Also, generating triplets using available batch algorithms is both time and space consuming. To address these challenges, we formulate the online Distance/Similarity learning problem using the robust Rescaled hinge loss function (Xu et al. 2017). Also, an efficient robust one-pass triplet construction algorithm is presented in this paper. We further extend our work by providing the low-rank version of proposed methods which not only reduces the computational cost significantly but also keeps the predictive performance of the learned metrics.

We study the effects of label noise in a DML task and conduct several experiments to measure the performance of the proposed methods at different noise levels. Extensive experimental results show that the proposed methods can effectively detect wrong label data and reduce their influences in DML tasks. Thus, they consistently outperform other peers online Distance/Similarity learning algorithms in noisy environments.

We intend to extend the work for online deep distance/similarity learning. Some other directions for future work are

I. Examining the performance of the proposed methods in other applications like *CBIR*.
II. Extension of the proposed methods in *imbalanced* environments.
III. Enhance the performance of the proposed online triplet construction algorithm.

## Acknowledgment

We would like to acknowledge the Machine Learning Lab in the Engineering Faculty of FUM for their kind and technical support.